\documentclass{article}

\usepackage{arxiv}

\usepackage[table,dvipsnames]{xcolor}
\usepackage{amsmath,amsfonts}
\usepackage{algorithmic}
\usepackage{algorithm}
\usepackage{array}
\usepackage[caption=false,font=normalsize,labelfont=sf,textfont=sf]{subfig}
\usepackage{textcomp}
\usepackage{stfloats}
\usepackage{url}
\usepackage{verbatim}
\usepackage{graphicx}
\usepackage{multicol }
\usepackage{cite}
\usepackage{natbib}
\usepackage[utf8]{inputenc} 
\usepackage[T1]{fontenc}    
\usepackage[pagebackref=true,breaklinks=true,colorlinks,bookmarks=false]{hyperref}
\usepackage{booktabs}       
\usepackage{nicefrac}       
\usepackage{microtype}      
\usepackage{enumitem}
\usepackage{caption}
\usepackage{multirow}
\usepackage{booktabs}
\usepackage{doi}

\usepackage[utf8]{inputenc}
\usepackage{soulutf8}
\usepackage{pifont}
\usepackage{amsmath}
\usepackage{amssymb}
\usepackage{breqn}
\usepackage{bm}
\usepackage{color,soul}
\usepackage[dvipsnames]{xcolor}
\usepackage{gensymb}

\usepackage{xcolor}
\hypersetup{
    colorlinks,
    linkcolor={red!80!black},
    citecolor={blue!50!black},
    urlcolor={blue!80!black}
}

\newcommand{\mytilde}{\raise.17ex\hbox{$\scriptstyle\mathtt{\sim}$}}
\DeclareMathOperator*{\argmin}{arg\,min}
\DeclareMathOperator*{\argmax}{arg\,max}

\let\emptyset\varnothing
\newcommand\given[1][]{\:#1\vert\:}

\def\mm{\mathbf{m}}

\def\xx{\mathbf{x}}

\def\MM{\mathbf{M}}

\def\dD{\mathcal{D}}

\def\lL{\mathcal{L}}

\def\nN{\mathcal{N}}

\def\xX{\mathcal{X}}
\def\yY{\mathcal{Y}}

\newcommand{\result}[1]{\ensuremath{#1}}

\newcommand{\tinyimagenet}{Tiny-ImageNet\textsubscript{1/2}}
\newcommand{\resultnof}[2]{#1 $\pm$ #2}

\title{Wake-Sleep Consolidated Learning}

\date{} 					

\author{ \href{https://orcid.org/0009-0001-8539-6299}{\includegraphics[scale=0.06]{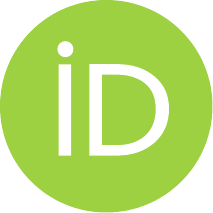}\hspace{1mm}Amelia Sorrenti}\\
        PeRCeiVe Lab \\
        University of Catania, Italy\\ 
        \And
        \href{https://orcid.org/0000-0002-1333-8348}{\includegraphics[scale=0.06]{orcid.pdf}\hspace{1mm} Giovanni Bellitto}\\
        PeRCeiVe Lab \\
        University of Catania, Italy\\ 
        \And
        \href{https://orcid.org/0000-0002-6122-4249}{\includegraphics[scale=0.06]{orcid.pdf}\hspace{1mm}Federica Proietto Salanitri}\\
        PeRCeiVe Lab \\
        University of Catania, Italy\\ 
        \And
        \href{https://orcid.org/0000-0002-6721-4383}{\includegraphics[scale=0.06]{orcid.pdf}\hspace{1mm}Matteo Pennisi}\\
        PeRCeiVe Lab \\
        University of Catania, Italy\\ 
        \And
        \href{https://orcid.org/0000-0002-2441-0982}{\includegraphics[scale=0.06]{orcid.pdf}\hspace{1mm}Simone Palazzo}\\
        PeRCeiVe Lab \\
        University of Catania, Italy\\ 
        \And
        \href{https://orcid.org/0000-0001-6653-2577}{\includegraphics[scale=0.06]{orcid.pdf}\hspace{1mm}Concetto Spampinato}\\
	PeRCeiVe Lab \\
        University of Catania, Italy\\ 
}

\renewcommand{\shorttitle}{Wake-Sleep Consolidated Learning}

\hypersetup{
pdftitle={A template for the arxiv style},
pdfsubject={q-bio.NC, q-bio.QM},
pdfauthor={David S.~Hippocampus, Elias D.~Striatum},
pdfkeywords={First keyword, Second keyword, More},
}

\begin{document}
\maketitle

\begin{abstract}
	We propose \textbf{Wake-Sleep Consolidated Learning (WSCL)}, a learning strategy leveraging Complementary Learning System theory and the wake-sleep phases of the human brain to improve the performance of deep neural networks for visual classification tasks in continual learning settings. 
Our method learns continually via the synchronization between distinct wake and sleep phases. During the wake phase, the model is exposed to sensory input and adapts its representations, ensuring stability through a dynamic parameter freezing mechanism and storing episodic memories in a short-term temporary memory (similarly to what happens in the hippocampus).
During the sleep phase, the training process is split into NREM and REM stages. In the NREM stage, the model's synaptic weights are consolidated using replayed samples from the short-term and long-term memory and the synaptic plasticity mechanism is activated, strengthening important connections and weakening unimportant ones. 
In the REM stage, the model is exposed to previously-unseen realistic visual sensory experience, and the dreaming process is activated, which enables the model to explore the potential feature space, thus preparing synapses to future knowledge.\\
We evaluate the effectiveness of our approach on three benchmark datasets: CIFAR-10, Tiny-ImageNet and FG-ImageNet. In all cases, our method outperforms the baselines and prior work, yielding a significant performance gain on continual visual classification tasks. Furthermore, we demonstrate the usefulness of all processing stages and the importance of dreaming to enable positive forward transfer. 

\end{abstract}


\section{Introduction}
Humans have a remarkable ability to continuously learn and retain past experiences while quickly adapting to new tasks and problems. On the contrary, machine learning has shown limitations when dealing with non-stationary data streams. This can be attributed to the inherent structure and optimization approaches of artificial neural networks, which differ significantly from how humans learn and build neural connectivity over a lifetime. The Complementary Learning Systems (CLS) theory~\cite{cls_2026,cls_1995} suggests that effective human learning occurs through the interplay of two learning processes originating from the hippocampus and neocortex brain regions. These regions interact to learn representations from experience (neocortex) while consolidating and sustaining long-term memory (hippocampus). This theory has inspired continual learning methods~\cite{pham2021dualnet,wang2022dualprompt} which translate CLS concepts into computational frameworks. DualNet~\cite{pham2021dualnet} employs two learning networks: a slow learner that emulates the memory consolidation process in the hippocampus and a fast learner that adapts current representations to new observations.
DualPrompt~\cite{wang2022dualprompt} addresses the challenge of adapting transformer models to new tasks while minimizing the loss of previous knowledge, using learnable prompts that are responsible for adapting to new data quickly, while preventing catastrophic forgetting. The specialization of prompt sets to their respective tasks is similar to how the hippocampus and neocortex specialize in complementary learning processes. DualNet and DualPrompt suggest that grounding artificial neural networks to cognitive neuroscience may result in improved performance, as they both achieve state-of-the-art performance on multiple benchmarks. Though promising, these approaches are rather rigid as the structures of the two learning parts (network architecture in DualNet; prompt format and positioning in DualPrompt) are defined \emph{a priori}, while neural networks in primates perform fast adaptation by flexibly re-configuring synapses while learning from new experience. 
Moreover, prior work does not consider the role of offline brain states such as sleep. Current theories suggest that sleep and dreaming play a crucial role in consolidating memories and facilitating learning, by increasing generalization of knowledge~\cite{pmid17173043,pmid15450165,shapiro_pnas}. During sleep, neurons are spontaneously active without external input and generate complex patterns of synchronized activity across brain regions~\cite{pmid8235588,pmid27849520}. This strong neural activity is believed to be due to the brain replaying and consolidating memories, while reorganizing synaptic connections.

In this work we propose Wake-Sleep Consolidated Learning (\textbf{WSCL}), extending the CLS theory by including wake-sleep states, in order to improve artificial neural networks' continual learning capabilities. This integration is achieved by introducing a sleep phase at training time that mimics the offline brain states during which synaptic connection, memory consolidation and dreaming occur. 

In WSCL, a deep neural network (DNN) is used to emulate the functions of the neocortex, while a two-layered buffer for short-term and long-term memory mimics the role of the hippocampus. Training is organized in two main phases: 1) a \emph{wake phase}, where fast adaption of the DNN to new sensory experience is carried out and episodic memories are stored in the short-term memory; 2) a \emph{sleep phase}, consisting of two alternating stages: a) Non-Rapid Eye Movement (NREM), where the network replays episodic memories collected during the wake step, consolidates past experiences in the long-term memory, and optimizes its neural connections to support synaptic plasticity; b) Rapid Eye Movement (REM), where dreaming simulates new experience, preparing the brain for future events. The hypothesis is that dreaming serves as an ``anticipatory'' mechanism, helping the brain to identify relationships between different types of information and making it easier to learn and remember new information.

Our computational formulation of the week-sleep process is tested on several benchmarks, including CIFAR-10, Tiny-ImageNet and FG-ImageNet. In all cases, our method outperforms the baselines and prior work, yielding a significant gain in classification tasks. Remarkably, WSCL approach is the first continual learning method yielding positive forward transfer, demonstrating its ability to prepare synapses to future knowledge.  We also show that all three steps are necessary: the wake stage is essential to ensure efficiency and to favor network plasticity by the NREM stage, while the REM stage helps to increase feature transferability and reduce the forgetting of acquired knowledge. 

\section{Related Work}
Continual Learning (CL)~\cite{de2019continual,parisi2019continual} is a branch of machine learning whose objective is to bridge the gap in incremental learning between humans and neural networks. McCloskey and Cohen~\cite{mccloskey1989catastrophic} highlight that the latter undergo \emph{catastrophic forgetting} of previously acquired knowledge in the presence of input distribution shifts. To mitigate this problem, several solutions have been proposed, introducing either adequate regularization terms~\cite{kirkpatrick2017overcoming, zenke2017continual}, specific architectural organization~\cite{schwarz2018progress,mallya2018packnet} or the rehearsal of a small number of previously encountered data points~\cite{robins1995catastrophic,rebuffi2017icarl,buzzega2020dark}.

While current solutions help reducing forgetting, real-world application proves difficult, as typical CL evaluations are carried out on unrealistic benchmarks~\cite{aljundi2019task,van2022three}. Most approaches tackling this challenging scenario combine a replay strategy~\cite{ratcliff1990connectionist,robins1995catastrophic,chaudhry2019tiny} to regularization on logits sampled throughout the optimization trajectory~\cite{buzzega2020dark}. Some works focus on memory management: GSS~\cite{aljundi2019gradient} introduces a specific optimization of the basic rehearsal formula meant to store maximally informative samples; HAL~\cite{chaudhry2021using} individuates synthetic replay data points that are maximally affected by forgetting. Other works propose tailored classification schemes: CoPE~\cite{De_Lange_2021_ICCV} uses class prototypes to ensure a gradual evolution of the shared latent space; ER-ACE~\cite{caccia2022new} makes the cross-entropy loss asymmetric to minimize imbalance between current and past tasks. 
Recent works introduce a surrogate optimization objective: CR~\cite{mai2021supervised} employs a supervised contrastive learning objective and OCM~\cite{guo2022online} leverages mutual information: both aim at learning features that are less subject to forgetting.

Our approach differs from these classes of methods, in that we take inspiration from cognitive neuroscience theory of learning (Complementary Learning Systems and wake-sleep) and exploits brain off-line states such as sleeping and dreaming. We demonstrate that alternating standard training with a revisited strategy that combines on-line and off-line stages makes the model more resilient to task shifts. 
Recently, a few neuroscience-informed CL methods have been proposed. Elastic Weight Consolidation (EWC)~\cite{kirkpatrick2017overcoming} and Synaptic Intelligence~\cite{zenke2017continual} employ regularization to preserve important weights learned during previous tasks while allowing the network to adapt to new tasks, emulating fast adaption happening in the neocortex. FearNet~\cite{kemker2017fearnet} adopts an auxiliary network (in line with CLS theory) to detect catastrophic forgetting and trigger knowledge-preserving regularization. Co2L~\cite{cha2021co2l} learns stable representations through contrastive learning and self-supervised distillation.

Two approaches similarly inspired by CLS theory are DualNet~\cite{pham2021dualnet} and DualPrompt~\cite{wang2022dualprompt}. DualNet employs two networks that loosely emulate slow and fast learning in humans. DualPrompt~\cite{wang2022dualprompt} also takes a cognitive approach, using learnable prompts to be paired to a pretrained transformer backbone. 
While both approaches yield good results, they ignore off-line states, that appear fundamental in human learning. 
Though not applied to continual learning yet, the wake-sleep algorithm has been shown to have the potential for learning improved and robust semantic representations~\cite{hintonWS,bornschein2015reweighted}.
Another related approach is Sleep Replay Consolidation~\cite{pmid36522325} that employs sleep-based training using local unsupervised Hebbian plasticity rules for mitigating catastrophic forgetting of ANN.

WSCL further unfolds the sleep phase by detailing the NREM and REM stages, integrating the dreaming process into the learning loop. This integration, which appears to contribute significantly to human learning, has a positive impact on the training of neural networks (as shown in the results).
The computational formulation of the wake-NREM-REM of WSCL is inspired by~\cite{pmid35384841}, where the role of adversarial dreaming for learning visual representations is preliminary investigated. However, simple strengthening of existing connections through unsupervised learning as proposed in~\cite{pmid36522325,pmid35384841} does not seem sufficient to build robust representations during sleep~\cite{shapiro_pnas}: our work thus explores more sophisticated restructuring of neural connections in the neocortex guided by the hippocampus.

\section{Method}
\begin{figure*}[htb!]
    \centering
    \includegraphics[width=1\textwidth]{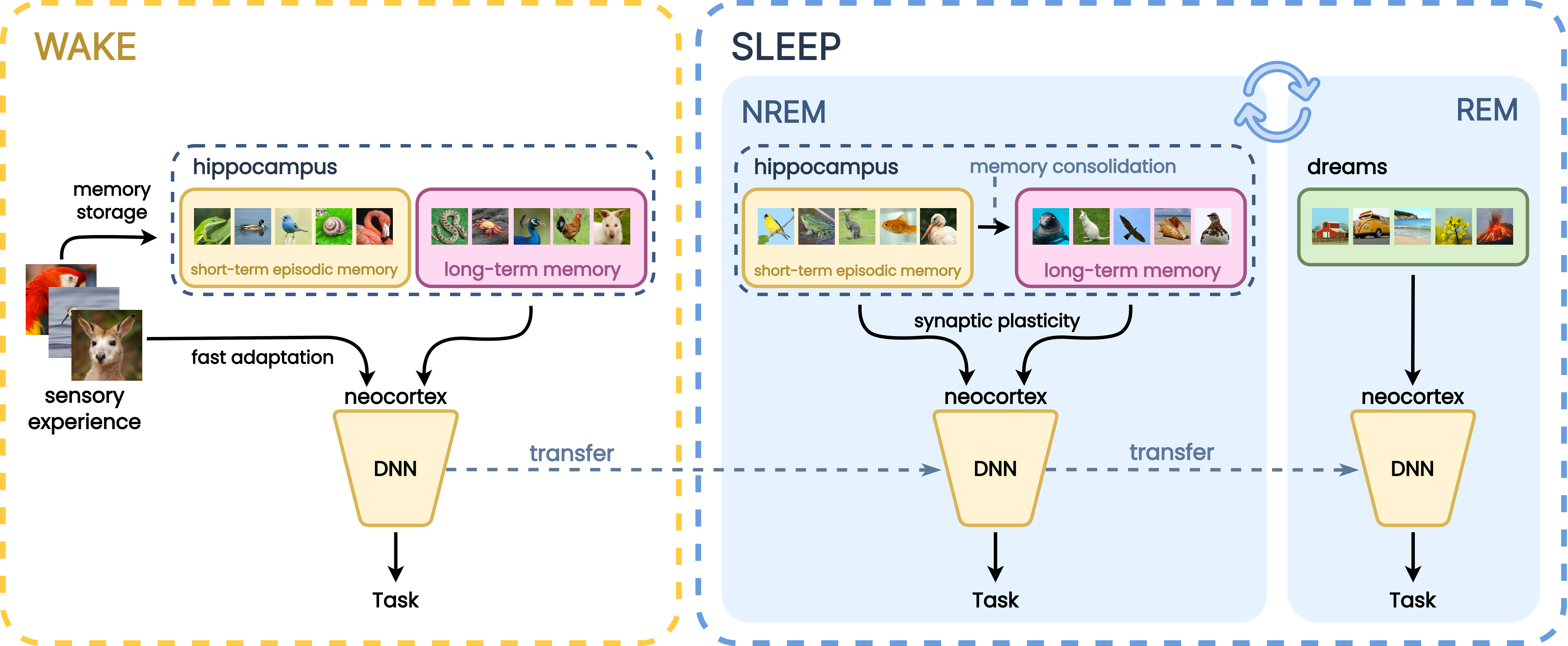}
    \caption{\textbf{Wake-Sleep Consolidated Learning}: in the wake stage, the model (which emulates the neocortex) fast adapts to the new sensory experience, storing episodic memories (as in the hippocampus) in the short-term memory to be replayed during sleep. The sleep phase foresees two alternating processes: 1) the NREM stage, where the DNN model consolidates its synapses based on the replayed (recent and past) samples and the long-term memory is updated; 2) the REM stage, where the DNN is trained with  dreamed samples to prepare the model for future sensory inputs.}
    \label{fig:overview}
\end{figure*}

An overview of the WSCL approach is presented in Fig.~\ref{fig:overview}, showing how the training stage on a new task is divided into two phases: a \emph{wake phase} and a \emph{sleep phase}. 

During the wake phase, the model is exposed to the new task, with the objective of performing fast adaptation of existing knowledge to the task characteristics. In this stage, the model quickly updates its parameters in order to find a balance between previously-acquired knowledge and new information, storing the latter in a short-term memory for later reuse during the sleep stage. In implementation terms, this balance is achieved by dynamically and adaptively freezing layer representations, identifying plasticity requirements for learning the new task while enforcing stability. Thus, during the wake stage, WSCL focuses primarily on learning general and transferable representation by combining both current and past experience.

In the sleep phase, the model consolidates newly acquired knowledge by revisiting the hippocampus short-term memory containing the task data, merging it into existing knowledge by updating synaptic connections, moving it into a long-term memory for future reference, and exploring the representational space through task-agnostic ``dreaming''. These stages are mapped into our training procedure by means of supervised training on task data, buffering task information in a (small) long-term memory, and employing an auxiliary dataset (uncorrelated to task data) as a surrogate for the generative process associated to dreaming.
Following the established literature, we pose continual learning as a supervised classification problem on a non-i.i.d. stream of data, with the assumption that \emph{task boundaries}, marking changes in the data distributions, are known at training time.
More formally, let $\dD =\{\dD_1, \dots, \dD_T\}$ be a sequence of data streams, where each pair $\left( \xx, y \right) \sim \dD_i$ denotes a data point $\xx \in \xX$ with the corresponding class label $y \in \yY$; the sample distributions (in terms of both the data point distribution and the class label distribution) of different $\dD_i$ and $\dD_j$ may vary --- for instance, class labels from $\dD_i$ might be different from those from $\dD_j$. 
Given a classifier $f: \xX \rightarrow \yY$, parameterized by $\bm{\theta}$,
the objective of continual learning is to train $f$ on $\mathcal{D}$, organized as a sequence of $T$ tasks $\{\tau_1, \dots, \tau_T\}$, under the constraint that, at a generic task $\tau_i$, the model receives inputs sampled from the corresponding data distribution only, i.e., $(\xx,y) \sim D_i$. The classification model may also keep a limited \emph{memory buffer} $\MM$ (assumed to be our long-term memory in the hippocampus) of past samples, to reduce forgetting of features from previous tasks. The model update step between tasks can be summarized as:
\begin{equation}
    \langle f, \bm{\theta}_{i-1}, \MM_{{{i-1}}}\rangle \xrightarrow{ \dD_{i}} \langle f, \bm{\theta}_{{i}}, \MM_{i}\rangle 
\end{equation}

where $\bm{\theta}_{i}$ and $\MM_{i}$ represent the set of model parameters and the memory buffer at the end of task $\tau_i$.

The training objective is to optimize a classification loss over the sequence of tasks (without losing accuracy on past tasks) by the model instance at the end of training:

\begin{dmath}
\argmin_{\bm{\theta}_T}
\sum_{i=1}^T
\mathbb{E}_{(\xx,y) \sim \dD_i}
\Bigl[
\lL \Bigl( f \left( \xx ; \bm{\theta}_T \right) , y \Bigr)
\Bigr]
\label{eq:cl_obj}
\end{dmath}
where $\lL$ is a generic classification loss (e.g., cross-entropy), which a continual learning model attempts to optimize while accounting for model \emph{plasticity} (the capability to learn current task data) and \emph{stability} (the capability to retain knowledge of previous tasks)~\cite{mccloskey1989catastrophic}. 

\subsection{Wake phase}
\label{sec:wake}

According to the established cognitive foundation, we define the waking stage in the proposed learning paradigm as the combination of two simultaneous processes, \emph{short-term memorization} and \emph{fast model adaptation}.

\noindent{\textbf{Short-term memorization}} has the objective of storing part of the current task experience, for later reuse --- in particular, for processing and consolidation during the sleep stage. In a continual learning setting, we model short-term memorization into $\MM_s$ as a sampling of task data $\dD_i$:
\begin{equation}
    \MM_s = \left\{ \left( \xx_j, y_j \right) \sim \dD_i \right\}_{j=1}^{N_s} ,
\end{equation}
where $N_s$ is the amount of samples collected from the $\dD_i$ distribution
\footnote{For brevity, we drop task index $i$ from short-term memory $\MM_s$, as it is re-created at each task.}.
Note that $\MM_s$ is reset during each wake phase and is distinguished from the \emph{long-term memory} $\MM_i$, which includes a smaller permanent number of samples $N_l$ from past tasks (in practice, the buffer of rehearsal-based methods). 

\noindent{\textbf{Fast model adaptation.}} In accordance to CLS theory~\cite{cls_2026,cls_1995}, we propose a method for fast model adaptation that employs parameter freezing during the wake stage to maximize stability and plasticity. Specifically, we propose to train the model for a limited number of iterations under varying parameter freezing settings, providing an opportunity to the model to rapidly learn new information in the wake stage while retaining the previous knowledge; in-depth consolidation of task information will be carried out separately in the sleep stage.
Unlike approaches such as DualNet, where the structure of the slow and fast networks are predefined, in WSCL the part of the network that reuses past knowledge and the part accounting for plasticity are identified on-line during the wake phase.

Formally, we want to model the joint probability between task data $\dD_i$, previous experience $\MM_{i-1}$, model parameters $\bm{\theta}_{i}$ and a binary freezing mask $\mm_i$, with the same dimensions as $\bm{\theta}_{i}$ and such that $m_{i,j}=1$ indicates that parameter $\theta_{i,j}$ should be frozen:
\begin{equation}
\label{eq:fma_joint}
P(\xx, y, \bm{\theta}_i, \mm_i) = P\left( y \given \xx, f\left(\xx, \bm{\theta}_i, \mm_i\right) \right)  P(\bm{\theta}_i, \mm_i) P(\xx) ,
\end{equation}
where $\xx$ and $y$ represent samples and labels from $\dD_i \cup \MM_{i-1}$. The first term of the decomposition of Eq.~\ref{eq:fma_joint} is the likelihood of correct labels given the input and the model prediction, while the joint distribution $P(\bm{\theta}_i, \mm_i)$ describes the relation between model parameters $\bm{\theta}_i$ and the freezing strategy defined by $\mm_i$. Assuming the independence between $\bm{\theta}_i$ and $\mm_i$, this distribution can be expressed as:
\begin{equation}
P(\bm{\theta}_i, \mm_i) = P(\bm{\theta}_i \given \mm_i) P(\mm_i) ,
\end{equation}
where
\begin{equation}
P(\bm{\theta}_i \given \mm_i) = \prod_j \nN \left( \theta_{i,j} ; \theta_{i-1,j}, \sigma_i^2) \right) ^ {1-m_{i,j}} .
\label{prior_w}
\end{equation}
In this formulation, we model the distribution of each parameter $\theta_{i,j}$ as a Gaussian distribution depending on the corresponding mask value $m_{i,j}$, which removes a term from the overall probability when $m_{i,j} = 1$. Note that the mean of each parameter is set to $\theta_{i-1,j}$, i.e., its value at the end of the previous task (or to 0 for the first task, based on common initialization strategies).

In order to model $P(\mm_i)$ in a practically feasible way, we employ some simplifying assumption based on the layered structure of deep learning models. Given $f = l_1 \circ l_2 \circ \dots \circ l_L$, where each $l_k$ represents a network layer with parameters $\bm{\theta}_{|k}$ and $\bm{\theta} = \left[ \bm{\theta}_{|1}, \dots, \bm{\theta}_{|L} \right]$, let us similarly define $\bm{0}_{|k}$ and $\bm{1}_{|k}$ as two tensors with the same size as $\bm{\theta}_{|k}$, with all values set to 0 and 1, respectively. Then, we impose that possible values for $\mm_i$ must be parameterized by a value $l$ as follows:
\begin{equation}
\mm_i(l) = \left[ \bm{1}_{|1}, \dots, \bm{1}_{|l}, \bm{0}_{|l+1}, \dots, \bm{0}_{|L} \right] \lor \mm_{i-1}
\end{equation}
with $l \in \left\{ 1, \dots, L \right\}$. In practice, parameters frozen at previous tasks must remain so at the current task, and a layer's parameters can only be frozen altogether if all previous layers are also frozen.

Given these constraints, our goal is to find the optimal binary mask $\mm_i$ that maximizes the likelihood of the labels $y$ given the inputs $\xx$ from current task $\dD_i$ and from long-term memory $\MM_{i-1}$.This is expressed as the following optimization problem:
\begin{equation}
\argmax_{\mm_i,\bm{\theta}_i} P\left( y \given \xx, f\left(\xx, \bm{\theta}_i, \mm_i\right) \right)  P(\bm{\theta}_i \given \mm_i) P(\mm_i) P(\xx)   
\end{equation}
where the optimization is over parameters $\bm{\theta}_i$ and all feasible binary masks $\mm_i$. Fast adaptation is thus carried out by maximizing this likelihood through the optimization of a loss function $\lL$:

\begin{equation}
\label{eq:fma_loss}
\begin{aligned}
    \lL_{\text{fma}} = & \mathbb{E}_{(\xx,y) \sim \dD_i} \left[
                    \lL \left( y, f\left(\xx, \bm{\theta}_i, \mm_i\right) \right)
                 \right] \\ 
                 & + \alpha \mathbb{E}_{(\xx,y) \sim \MM_{i-1}} \left[ \lL \left( y, f\left(\xx, \bm{\theta}_i, \mm_i\right) \right) \right] ,
\end{aligned}
\end{equation}

where $\mm_i$ varies as described above, and $\alpha$ is a weighing factor between data sources. It is important to notice that, while optimizing for $\mm_i$ necessarily requires updating $\bm{\theta}_i$ as well (since freezing, per se, does not alter inference performance), the objective is to prepare the model by identifying the optimal set of parameters that should be kept from previous tasks in a way that ensures both knowledge retainment and room for plasticity. For this reason, optimization is carried out for a single epoch over $\dD_i$. Note that the choice of $\lL$ is arbitrary: the proposed formulation allows for plugging in any existing continual learning method, enhancing it with the proposed training strategy.  

\subsection{Sleep phase}

During sleep, the brain cycles multiple times through two phases, known as rapid eye movement (REM) and non-rapid eye movement (NREM) sleep. In the NREM phase, the hippocampus replays and consolidates the information acquired at waking time by facilitating its transfer to the neocortex, where long-term memory storage occurs~\cite{pmid12590838,pmid17173043}. 
REM sleep is thought to play a role in creativity and problem-solving~\cite{pmid26779078,Schwartz2003}, allowing the brain to form new connections and generate novel ideas. 
In our WSCL approach, we analogously distinguish between two alternating training modalities, conceptually mapped to the NREM and REM phases. 
During the former, we access examples from the current task (stored in the short-term memory) and from previous tasks (retrieved from long-term memory) to train the model --- partially frozen during the wake stage --- and stabilizing present knowledge. In the REM stage, we emulate the dreaming process by providing the model with examples from an external data source, with classes unrelated to any continual learning task. This approach allows the model to learn task-agnostic features which can be interpreted as a prior knowledge supporting task-specific learning and forward transfer. 

\noindent{\textbf{NREM stage.}} The main objective of this stage is to transfer information from the short-term memory $\MM_s$, built in the precedent wake phase, to the model, strengthening the synaptic connections associated to the current task and thus enforcing plasticity, while retaining previously acquired knowledge thanks to long-term memory $\MM_{i-1}$. In this setting, we apply parameter freezing mask $\mm_i$ (defined in the wake phase), which is however not  updated in the process.

Formally, in this stage we model the same distribution as in Eq.~\ref{eq:fma_joint}, but optimize for $\bm{\theta}_i$ alone, while leaving $\mm_i$ constant. The objective thus becomes:
\begin{equation}
\label{eq:argmax_nrem}
\argmax_{\bm{\theta}_i} P\left( y \given \xx, f\left(\xx, \bm{\theta}_i, \mm_i\right) \right)  P(\bm{\theta}_i \given \mm_i) P(\xx) ,
\end{equation}
where the prior on parameters $P\left(\bm{\theta}_i \given \mm_i \right)$ is essentially the same as in Eq.~\ref{prior_w}, with the difference that the mean of the distribution is the value of $\bm{\theta}_i$ as computed at the end of the wake stage, rather than $\bm{\theta}_{i-1}$. Optimizing the above objective amounts to minimizing a variant of the loss in Eq.~\ref{eq:fma_loss}:

\begin{equation}
\begin{aligned}
\label{eq:nrem_loss}
\lL_{\text{NREM}} = & \mathbb{E}_{(\xx,y) \sim \MM_s} \left[
                    \lL \left( y, f\left(\xx, \bm{\theta}_i, \mm_i\right) \right)
                 \right]\\
& + \alpha \mathbb{E}_{(\xx,y) \sim \MM_{i-1}} \left[ \lL \left( y, f\left(\xx, \bm{\theta}_i, \mm_i\right) \right) \right],
\end{aligned}
\end{equation}

where $\MM_s$ is employed instead of the whole dataset $\dD_i$.

In this stage, we also gradually update long-term memory $\MM_i$, using reservoir sampling~\cite{lopez2017gradient} to inject task experience from short-term memory $\MM_s$ into $\MM_i$, so that it becomes available to future tasks.

\noindent{\textbf{REM stage.}} We approximate the sleeping mechanism performed by the human brain in the REM stage by providing the model 
with an additional source of previously unseen knowledge (a ``dreaming'' dataset with no semantic overlap with CL classes), that can help the model to generalize better to new and unseen data, as suggested by cognitive literature~\cite{pmid35384841}. Please note that the dreaming dataset can be replaced with a generative model, either BigGAN~\cite{brock2018large} or a diffusion model~\cite{Rombach_2022_CVPR}, without losing generalization. Indeed, we did not note any significant performance change when replacing the dreaming dataset with a GAN trained on it, as also observed in~\cite{bellitto2022effects}.
\newcommand{\daux}{\ensuremath{\dD_{\text{dream}}}}
\newcommand{\yaux}{\ensuremath{\yY_{\text{dream}}}}

Let ${\daux}$ be the dreaming dataset from which we can sample data points $\left( \xx, y \right) \sim \daux$, with $\xx \in \xX$ and class label $y \in \yaux$. We assume that $\yaux \cap \yY = \emptyset$ (the latter being the set of continual learning classes), to prevent any overlap between auxiliary and continual learning classes. Given this premise, the proposed optimization objective becomes:
\begin{equation}
\argmax_{\bm{\theta}_i} P\left( y \given \xx, f\left(\xx, \bm{\theta}_i, \mm_i\right) \right)  P(\bm{\theta}_i \given \mm_i) P(\xx) ,
\end{equation}
where $(\xx, y) \sim \daux$, while the other terms are the same as in Eq.~\ref{eq:argmax_nrem}. This objective is then mapped to a training loss function defined as:
\begin{equation}
\label{eq:rem_loss}
\lL_{\text{REM}} = \mathbb{E}_{(\xx,y) \sim \daux} \left[
                    \lL \left( y, f\left(\xx, \bm{\theta}_i, \mm_i\right) \right)
                 \right] .
\end{equation}
During REM stage, training with two distinct class label sets, $\yY$ from the continual learning problem and $\yaux$ from the dreaming dataset has been addressed following the procedure reported in~\cite{bellitto2022effects}.

\section{Experimental Evaluation}
\subsection{Benchmarks}
We test WSCL on several continual learning benchmarks obtained by taking image classification datasets and splitting their classes equally into a series of disjoint tasks. Moreover, since REM stage requires additional dreaming samples, for each benchmark we also identify its dreamed-counterpart:
\begin{itemize}[noitemsep,nolistsep,leftmargin=*]
    \item \textbf{Split CIFAR-10}~\cite{zenke2017continual}, a widely-used image classification dataset obtained by splitting CIFAR-10 images into 5 binary classification tasks. Its counterpart used for the REM stage consists of a subset of 50 CIFAR-100 classes, selected after removing those with semantic relations to CIFAR-10.
    \item \textbf{Split FG-ImageNet}\footnote{Split\ FG-ImageNet is derived from \url{https://www.kaggle.com/datasets/ambityga/imagenet100}} is a fine-grained image classification benchmark with 100 classes of animals, used to test CL methods on a more challenging task.
    The dreaming counterpart consists of additional 100 classes taken from ImageNet, after removing all synsets derived from ``organism''. 
    \item \textbf{Tiny-ImageNet}~\cite{Le2015TinyIV} is a subset of ImageNet consisting of 200 classes with 500 images each, resized to 64$\times$64. We employ the first 100 classes as the main training dataset \emph{\tinyimagenet} (organized as 5 tasks of 20 classes) and the remaining 100 classes as the dreaming dataset. 
    \end{itemize}
\subsection{Training procedure}
Our approach employs a ResNet-18 backbone for feature extraction and classification. ResNet-18 includes, at a high level, four \emph{layers} with two {blocks} each, for a total of eight blocks\footnote{\url{https://pytorch.org/vision/master/_modules/torchvision/models/resnet.html}} With reference to the definition of model $f$ in Sect.~\ref{sec:wake}, we map each layer $l_i$ to each ResNet-18 block.

In the wake stage of task $i$, we train multiple instances of the model, starting from parameters $\bm{\theta}_i$, with all possible configurations of $\mm_i$: if the deepest frozen layer is $l_j$, the number of possible values for $\mm_i$ is $L - j + 1$, with $L$ being the total number of layers. Training is carried out for a single epoch with mini-batch SGD and a learning rate of 0.03.
Batch size is set to 32 for CIFAR-10 and \tinyimagenet, and to 8 for FG-ImageNet. The $\alpha$ hyperparameter in Eq.~\ref{eq:fma_loss} is set to 1, and the $N_s$ dimension of the short-term buffer to 5,000.
It is important to mention that, in our implementation, the optimization of Eq.~\ref{eq:fma_loss} (fast model adaptation loss $\lL_{\text{fma}}$) and Eq.~\ref{eq:nrem_loss} (NREM loss $\lL_{\text{NREM}}$) on long-term memory $\MM_i$ is carried out on disjoint portions of the whole set of stored samples. In particular, 10\% of $\MM_i$ is used when optimizing $\lL_{\text{fma}}$, while the remaining 90\% is used for $\lL_{\text{NREM}}$. This separation mitigates the risk of overfitting of $\lL_{\text{NREM}}$ on data that will be used, in the wake phase, to determine to which extent model layers should be frozen: indeed, in case of overfitting, the wake phase would encourage model freezing, as it would more easily minimize the corresponding loss term.\\
In the sleep stage, we train the model using $\lL_{\text{NREM}}$ and the $\lL_{\text{REM}}$ losses at alternately batches. We perform 10 epochs of training, with the same optimizer settings and hyperparameters as above.\\
All the reported results are computed in the class-incremental setting.

\label{sec:training}
\subsection{Results}
\label{sec:results}
\begin{table*}[ht!]
\caption{\textbf{Class-incremental final average accuracy (\textbf{FAA}) and forward transfer (\textbf{FWT}) of rehearsal-based methods}, with and without WSCL, for different buffer sizes.}
\label{tab:results}
\centering
\renewcommand{\arraystretch}{1.2}
\resizebox{.99\linewidth}{!}{%
\begin{tabular}{l|rrrrrrrrrrrr} 
\toprule

\textbf{Method}                                      &\multicolumn{4}{c}{\textbf{CIFAR-10}}                                     & \multicolumn{4}{c}{\textbf{\tinyimagenet}}                               &\multicolumn{4}{c}{\textbf{FG-ImageNet}}\\ 
\cmidrule(lr){1-1} \cmidrule(lr){2-5} \cmidrule(lr){6-9} \cmidrule(lr){10-13}   
\arrayrulecolor{black}
Joint                                               & \multicolumn{4}{c}{\result{85.15}}                                        & \multicolumn{4}{c}{\result{50.81}}                                       & \multicolumn{4}{c}{\result{43.39}}\\
Fine-tune                                           & \multicolumn{4}{c}{\result{19.47}}                                        & \multicolumn{4}{c}{\result{13.84}}                                       & \multicolumn{4}{c}{\result{3.88}}\\
\cmidrule(lr){1-1} \cmidrule(lr){2-5} \cmidrule(lr){6-9} \cmidrule(lr){10-13}  
\textbf{Buffer size}                                & \multicolumn{2}{c}{\textbf{200}}     & \multicolumn{2}{c}{\textbf{500}}   & \multicolumn{2}{c}{\textbf{200}}    & \multicolumn{2}{c}{\textbf{500}}   & \multicolumn{2}{c}{\textbf{200}}  & \multicolumn{2}{c}{\textbf{1000}}\\ 
\cmidrule(lr){1-1} \cmidrule(lr){2-3} \cmidrule(lr){4-5} \cmidrule(lr){6-7} \cmidrule(lr){8-9} \cmidrule(lr){10-11} \cmidrule(lr){12-13}
                                                    & \textbf{FAA}      & \textbf{FWT}     & \textbf{FAA}    & \textbf{FWT}     & \textbf{FAA}     & \textbf{FWT}     & \textbf{FAA}    & \textbf{FWT}       & \textbf{FAA}   & \textbf{FWT}     & \textbf{FAA}     & \textbf{FWT}     \\
ER~\cite{caccia2022new}                             & \result{48.76}    & \result{-7.36}   & \result{59.75}  & \result{-12.20}  & \result{16.25}   & \result{-1.00}   & \result{21.07}  & \result{-1.32}     & \result{4.23}  & \result{-1.05}   & \result{5.05}  & \result{-1.02}\\
\rowcolor{gray!10}
\hspace{0.1 cm } {$\hookrightarrow$\textbf{WSCL}}   & \result{\textbf{51.86}}   & \result{1.68}   & \result{\textbf{63.71}}  & \result{6.03}    & \result{\textbf{18.81}}   & \result{12.41}   & \result{\textbf{23.63}}  & \result{12.60}     & \result{\textbf{6.01}}  & \result{3.82}    & \result{\textbf{15.26}}    & \result{3.17}\\
DER++~\cite{buzzega2020dark}                        & \result{57.35}    & \result{-12.29}  & \result{69.06}  & \result{-6.23}   & \result{16.62}   & \result{-0.84}   & \result{23.40}  & \result{-1.06}     & \result{5.95}  & \result{-0.08}   & \result{8.59}     & \result{-1.05}\\
\rowcolor{gray!10}
\hspace{0.1 cm } {$\hookrightarrow$\textbf{WSCL}}   & \result{\textbf{63.97}}    & \result{1.06}    & \result{\textbf{72.33}}  & \result{2.83}    & \result{\textbf{23.70}}   & \result{12.16}   & \result{\textbf{31.81}}  & \result{12.24}     & \result{\textbf{6.48}}  & \result{1.78}    & \result{\textbf{11.70}}    & \result{2.31}\\
\arrayrulecolor{gray}
\arrayrulecolor{black}
ER-ACE~\cite{chaudhry2019tiny}                      & \result{59.98}    & \result{-8.58}   & \result{67.17}  & \result{-8.97}   & \result{27.81}   & \result{-0.73}   & \result{32.10}  & \result{-0.94}     & \result{9.42}  & \result{-1.04}    & \result{11.58}    & \result{-1.17}\\
\rowcolor{gray!10}
\hspace{0.1 cm } {$\hookrightarrow$\textbf{WSCL}}   & \result{\textbf{71.15}}   & \result{0.48}    & \result{\textbf{74.18}}  & \result{-1.87}   & \result{\textbf{35.68}}   & \result{8.60}    & \result{\textbf{41.25}}  & \result{9.06}      & \result{\textbf{12.51}} &\result{1.83}      & \result{\textbf{20.51}}    & \result{1.19}\\ 
\arrayrulecolor{black}
\midrule
\textbf{DualNet}~\cite{pham2021dualnet}             & \result{31.31}   & \result{-5.89}     & \result{43.20} & \result{-7.41}    & \result{16.45}   & \result{-0.78}    & \result{18.98} & \result{-0.96}       & \result{9.78}  &\result{-1.03}           & \result{16.54} & \result{-1.96}\\
\textbf{CoPE}~\cite{De_Lange_2021_ICCV}             & \result{21.20}   & \result{-3.63}     & \result{23.64} & \result{-4.23}     & \result{16.50}   & \result{-0.87}    & \result{20.50} & \result{-1.05}         & \result{6.23}   &\result{-0.98}           & \result{12.57}   & \result{-1.23}\\

\bottomrule
\end{tabular}}
\end{table*}

We first evaluate how WSCL contributes to classification accuracy of state-of-the-art models. To accomplish this, we select recent rehearsal-based methods, namely, DER++~\cite{buzzega2020dark}, ER-ACE~\cite{caccia2022new} and ER~\cite{chaudhry2019tiny}, and compare their performance when the WSCL training strategy is employed, by plugging them in as the $\lL$ loss term in Eq.~\ref{eq:fma_loss}, \ref{eq:nrem_loss}, \ref{eq:rem_loss}. We address rehearsal-based methods only, as WSCL requires a memory buffer to model long-term memory. 
We report \emph{final average accuracy (FAA)} after training on the last task in the class-incremental setting\footnote{Task-incremental results are available in the supplementary material.}.
We further provide a lower bound, consisting of training without any countermeasure to forgetting (\emph{Fine-tune}), and an upper bound given by training all tasks jointly (\emph{Joint}).

Results in Table~\ref{tab:results} show that, on all three benchmarks, WSCL leads to a significant performance gain that varies from about 2 percent points on FG-ImageNet to 12 percent points on CIFAR-10, substantiating our claims on the importance of leveraging human learning strategies for building better computational methods. 
Table~\ref{tab:results} also reports the comparison with: a) DualNet~\cite{pham2021dualnet}, which leverages CLS theory and the same backbone, i.e., ResNet-18; b) CoPE~\cite{De_Lange_2021_ICCV} that integrates contrastive learning --- another technique inspired by cognitive neuroscience~\cite{pmid35384841} ---  for better feature transferability to later tasks\footnote{Results for DualNet and CoPE are computed using their original implementations and hyperparameters.}. 
We do not include DualPrompt~\cite{wang2022dualprompt} as it uses a large pre-trained ViT~\cite{dosovitskiy2021image} as a backbone, leading to an unfair comparison with the simpler ResNet-18. 
All methods combined with our WSCL strategy improve over DualNet (up to about 40 percent points) and CoPE, demonstrating how mimicking human learning more strictly improves performance even in a purely discriminative supervised learning regime. 

\begin{table*}[t!]
\caption{\textbf{Forgetting Class-IL}}
\label{tab:forgetting}
\centering
\renewcommand{\arraystretch}{1.2}
\begin{tabular}
{>{\hspace{0pt}}m{0.13\linewidth}|   >{\centering\hspace{0pt}}m{0.095\linewidth}  >{\centering\hspace{0pt}}m{0.095\linewidth}| >{\centering\hspace{0pt}}m{0.095\linewidth}>{\centering\hspace{0pt}}m{0.095\linewidth}|    >{\centering\hspace{0pt}}m{0.095\linewidth}  >{\centering\arraybackslash\hspace{0pt}}m{0.095\linewidth}}

\toprule
\textbf{Method}                                      &\multicolumn{2}{c}{\textbf{CIFAR-10}}                 & \multicolumn{2}{c}{\textbf{\tinyimagenet}}      &\multicolumn{2}{c}{\textbf{FG-ImageNet}}\\   
\midrule
& \multicolumn{6}{c}{\emph{\textbf{Class-incremental learning}}} \\
\midrule
\textbf{Buffer size}                                & \multicolumn{1}{c}{\textbf{200}}          & \multicolumn{1}{c}{\textbf{500}}                  & \multicolumn{1}{c}{\textbf{200}}       & \multicolumn{1}{c}{\textbf{500}}                   & \multicolumn{1}{c}{\textbf{200}}        & \multicolumn{1}{c}{\textbf{1000}}\\
\cmidrule(lr){1-1} \cmidrule(lr){2-3} \cmidrule(lr){4-5} \cmidrule(lr){6-7}
ER                             & \result{56.66}    & \result{43.21}    & \result{62.63}    & \result{58.16}    & \result{74.04}    & \result{73.45}\\
\rowcolor{gray!10}
\hspace{0.1 cm} {$\hookrightarrow$\textbf{WSCL}}   & \result{50.23}    & \result{36.04}    & \result{56.71}    & \result{50.63}    & \result{76.79}    & \result{63.93}\\
DER++                          & \result{31.23}    &\result{22.63}    &\result{62.15}    & \result{50.81}    & \result{67.10}    & \result{63.63}\\
\rowcolor{gray!10}
\hspace{0.1 cm} {$\hookrightarrow$\textbf{WSCL}}   & \result{35.53}    & \result{23.52}    & \result{51.30}    & \result{43.91}    & \result{59.84}    & \result{52.39}\\
ER-ACE                         & \result{16.55}    & \result{15.21}    & \result{34.41}    & \result{28.15}    & \result{32.61}    & \result{36.44}\\
\rowcolor{gray!10}
\hspace{0.1 cm} {$\hookrightarrow$\textbf{WSCL}}   & \result{11.78}    & \result{10.69}    & \result{28.23}    & \result{23.29}    & \result{27.24}    & \result{33.53}\\
\bottomrule
\end{tabular}
\end{table*}

We also measure \emph{forward transfer (FWT)}, a desirable property in CL that indicates how much a model leverages previous knowledge to learn a new task~\cite{lopez2017gradient}.
Forward transfer is estimated as the average difference between the accuracy of a task when learning it in a CL setting and when learning it from random initialization (details in~\cite{lopez2017gradient}). 
Table~\ref{tab:results} shows how WSCL tends to enhance FWT, turning it from negative to positive values. This is highly remarkable as the majority of existing CL methods show a negative forward transfer.

Furthermore, it is equally important to measure forgetting (the lower, the better) to assess how well an approach tackles no-iid data. Cross-checking results in Table \ref{tab:forgetting} with those available in Table~\ref{tab:results} highlights how WSCL effectively reduces forgetting, while enhancing  forward transfer skills and accuracy performance in a way sensibly higher than the baselines.

We further expand the comparison by grounding WSCL to other prominent continual learning methods\footnote{Results obtained using the original code released along with the relative papers.}.
For this evaluation, we employ of model that yields highest results, i.e., ER-ACE combined to WSCL (as shown in Table~\ref{tab:results}). 
As shown in Table \ref{tab:competitors}, ER-ACE~w/~WSCL (indicated as "ours") significantly outperforms all existing methods in both class-incremental and task-incremental settings. 
Notably, when excluding the buffer for training ER-ACE w/ WSCL (which means using the model without NREM stage, indicated in Table~\ref{tab:competitors} as Wake+REM), it achieves a substantial performance improvement, from approximately 12\% (on \tinyimagenet) to about 23\% on CIFAR-10, over 
existing buffer-free methods, namely, LwF~\cite{li2017learning}, SI~\cite{zenke2017continual}, and oEWC~\cite{schwarz2018progress}.

\begin{table*}[h!]
\centering
\caption{\textbf{Comparison with SOTA methods}, for different buffer sizes.}
\label{tab:competitors}
\rowcolors{15}{gray!10}{white}
\renewcommand{\arraystretch}{1.2}
\resizebox{.9\linewidth}{!}{%
\begin{tabular}{l|cc|cc|cc}
\toprule
\textbf{Method}                                      &\multicolumn{2}{c}{\textbf{CIFAR-10}}                                         & \multicolumn{2}{c}{\textbf{\tinyimagenet}}                                 &\multicolumn{2}{c}{\textbf{FG-ImageNet}}\\
\midrule
Joint                                               & \multicolumn{2}{c}{\resultnof{85.15}{1.99}}                                   & \multicolumn{2}{c}{\resultnof{50.81}{1.65}}                                & \multicolumn{2}{c}{\resultnof{43.39}{1.76}}\\
\rowcolor{gray!10}
Fine-tune                                           & \multicolumn{2}{c}{\resultnof{19.47}{0.10}}                                   & \multicolumn{2}{c}{\resultnof{13.84}{0.55}}                                & \multicolumn{2}{c}{\resultnof{3.88}{0.33}}\\ 
\midrule
\multicolumn{7}{c}{\textbf{Buffer-free methods}}\\
\midrule
LwF~\cite{li2017learning}                           & \multicolumn{2}{c}{\resultnof{19.33}{0.16}}                                   & \multicolumn{2}{c}{\resultnof{13.87}{1.11}}                                & \multicolumn{2}{c}{\resultnof{3.83}{0.11}}\\
\rowcolor{gray!10}
SI~\cite{zenke2017continual}                        & \multicolumn{2}{c}{\resultnof{19.27}{0.23}}                                   & \multicolumn{2}{c}{\resultnof{13.12}{1.63}}                                & \multicolumn{2}{c}{\resultnof{3.75}{0.35}}\\
oEWC~\cite{schwarz2018progress}                     & \multicolumn{2}{c}{\resultnof{18.96}{0.24}}                                   & \multicolumn{2}{c}{\resultnof{13.87}{0.53}}                                & \multicolumn{2}{c}{\resultnof{3.48}{0.18}}\\
\midrule
\rowcolor{gray!10}
\textbf{Ours (Wake+REM)}                            & \multicolumn{2}{c}{\resultnof{\textbf{41.58}}{3.94}}                          & \multicolumn{2}{c}{\resultnof{\textbf{25.68}}{0.44}}                       & \multicolumn{2}{c}{\resultnof{\textbf{6.27}}{0.89}}\\
\midrule  
\multicolumn{7}{c}{\textbf{Buffer-based methods}}\\
\midrule  
\textbf{Buffer size}                                & \multicolumn{1}{c}{\textbf{200}}    & \multicolumn{1}{c}{\textbf{500}}        & \multicolumn{1}{c}{\textbf{200}}  & \multicolumn{1}{c}{\textbf{500}}       & \multicolumn{1}{c}{\textbf{200}}    & \multicolumn{1}{c}{\textbf{1000}}\\

\cmidrule(lr){1-1} \cmidrule(lr){2-3} \cmidrule(lr){4-5} \cmidrule(lr){6-7}

ER~\cite{caccia2022new}                             & \resultnof{48.76}{0.57}             & \resultnof{59.75}{2.51}                 & \resultnof{16.25}{0.85}           & \resultnof{21.07}{1.43}                & \resultnof{4.23}{0.15}              & \resultnof{5.05}{0.51}\\
DER++~\cite{buzzega2020dark}                        & \resultnof{57.35}{5.47}             & \resultnof{69.06}{1.24}                 & \resultnof{16.62}{1.76}           & \resultnof{23.40}{1.66}                & \resultnof{5.95}{0.49}              & \resultnof{8.59}{1.11}\\
ER-ACE~\cite{chaudhry2019tiny}                      & \resultnof{59.98}{2.65}             & \resultnof{67.17}{1.54}                 & \resultnof{27.81}{1.24}           & \resultnof{32.10}{2.21}                & \resultnof{9.42}{0.78}              & \resultnof{11.58}{3.59}\\
A-GEM~\cite{chaudhry2019efficient}                  & \resultnof{19.45}{0.25}             & \resultnof{20.21}{0.38}                 & \resultnof{13.75}{0.37}           & \resultnof{13.56}{0.39}                & \resultnof{4.00}{0.20}              & \resultnof{4.15}{0.06}\\
BiC~\cite{wu2019large}                              & \resultnof{55.03}{1.93}             & \resultnof{66.24}{1.65}                 & \resultnof{16.26}{0.87}           & \resultnof{12.88}{5.50}                & \resultnof{8.10}{2.75}              & \resultnof{7.03}{4.71}\\
FDR~\cite{benjamin2018measuring}                    & \resultnof{38.72}{8.93}             & \resultnof{31.91}{5.08}                 & \resultnof{17.67}{1.04}           & \resultnof{23.17}{1.69}                & \resultnof{4.44}{0.77}              & \resultnof{3.91}{0.22}\\
GEM~\cite{lopez2017gradient}                        & \resultnof{21.93}{2.04}             & \resultnof{20.80}{0.23}                 & \resultnof{14.57}{0.57}           & \resultnof{15.20}{1.28}                & \resultnof{4.36}{0.11}              & \resultnof{4.29}{0.28}\\
GDumb~\cite{prabhu2020gdumb}                        & \resultnof{33.81}{1.52}             & \resultnof{46.01}{3.26}                 & \resultnof{5.74}{0.45}            & \resultnof{9.85}{0.50}                 & \resultnof{4.54}{0.23}              & \resultnof{9.79}{1.18}\\
GSS~\cite{aljundi2019gradient}                      & \resultnof{41.36}{6.46}             & \resultnof{48.83}{4.41}                 & \resultnof{15.92}{0.88}           & \resultnof{18.15}{0.61}                & \resultnof{4.05}{0.42}              & \resultnof{4.46}{1.20}\\
iCaRL~\cite{rebuffi2017icarl}                       & \resultnof{64.52}{1.18}             & \resultnof{60.94}{1.34}                 & \resultnof{20.40}{0.36}           & \resultnof{22.68}{0.30}                & \resultnof{10.40}{0.20}             & \resultnof{11.17}{0.79}\\
LUCIR~\cite{hou2019learning}                        & \resultnof{53.48}{7.62}             & \resultnof{63.01}{3.40}                 & \resultnof{22.65}{1.18}           & \resultnof{32.15}{0.88}                & \resultnof{6.08}{0.32}              & \resultnof{13.19}{0.32}\\
RPC~\cite{pernici2021class}                         & \resultnof{49.37}{1.47}             & \resultnof{55.19}{2.73}                 & \resultnof{16.58}{0.52}           & \resultnof{20.95}{0.59}                & \resultnof{4.13}{0.16}              & \resultnof{5.83}{0.30}\\

\midrule
\textbf{Ours}                       & \resultnof{\textbf{71.15}}{2.15}    & \resultnof{\textbf{74.18}}{1.28}        & \resultnof{\textbf{35.68}}{1.18}  & \resultnof{\textbf{41.25}}{1.75}       & \resultnof{\textbf{12.51}}{0.86}    & \resultnof{\textbf{20.51}}{0.56}\\ 
\arrayrulecolor{black}

\arrayrulecolor{black}
\bottomrule
\end{tabular}}
\end{table*}
\subsection{Model Analysis}
\begin{figure*}[htb!]
\centering
    \includegraphics[width=0.9\textwidth]{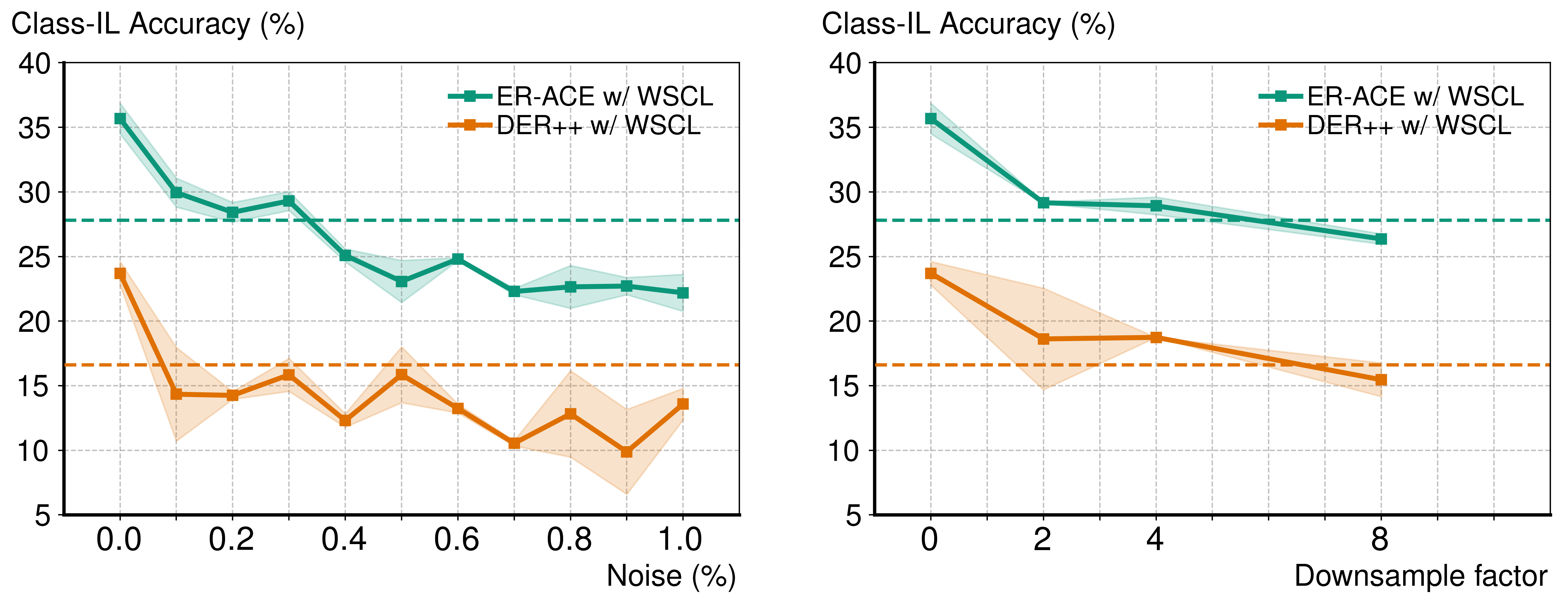}
    \caption{\textbf{Impact of dreaming quality}, in terms of noise (left) and image resolution (right). Results refer to ER-ACE and DER++ with WSCL (solid lines) and without it (dotted line).}
    \label{fig:ablation_noise}
\end{figure*}

\begin{figure}[ht!]
    \centering
    \includegraphics[width =.7\linewidth]{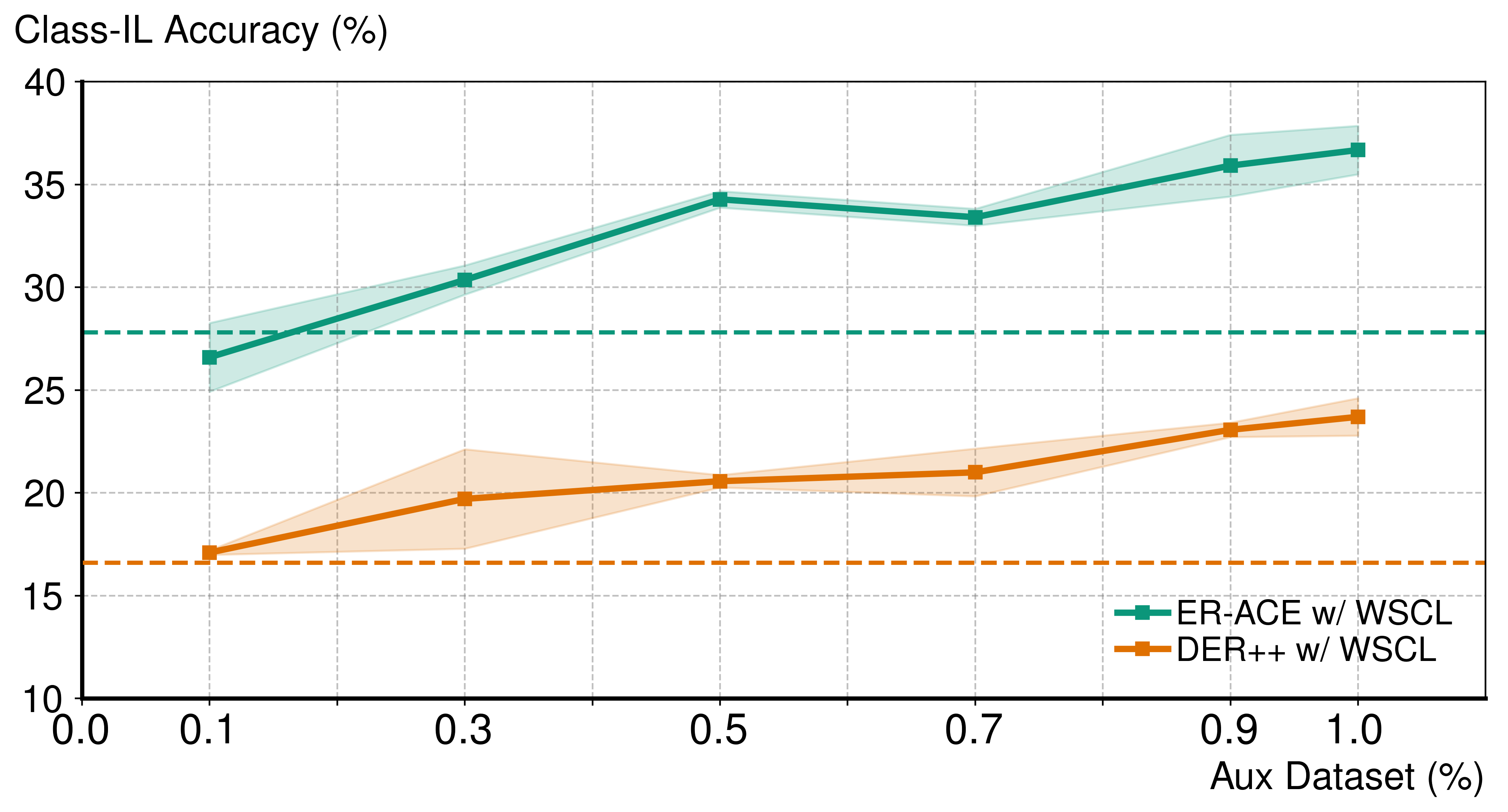}
    \caption{\textbf{Impact of dreaming dataset dimension}. Results refer to ER-ACE and DER++ with WSCL (solid lines) and without it (dotted line).}
    \label{fig:dim_dreaming}
\end{figure}

Model analysis is mainly carried out using ER-ACE (the best-performing method from Table~\ref{tab:results}) as baseline, on the \tinyimagenet dataset.
We first ablate the processing phases of WSCL: results in Table~\ref{tab:res_ablation} show how the NREM/REM sleep states equally contribute to the final model performance.
Interestingly, the REM phase is responsible for positive forward transfer, which is consistent with cognitive neuroscience evidence that REM prepares brain synapses to future experience~\cite{pmid26779078,Schwartz2003}.

We then evaluate the impact of the quality of dreaming, by adding Gaussian noise (at different percentages) and reducing the spatial resolution of dreaming samples. Fig.~\ref{fig:ablation_noise} indicates that WSCL still outperforms the baseline when dreaming images are affected by noise up to 30\% or scaled down by $6\times$, suggesting that the role of REM stage in consolidating knowledge is mostly independent from the visual details of the dreamed samples, which merely serve to learn additional reusable features.

\begin{table}[ht!]
\centering
\caption{\textbf{Ablation on the WSCL processing stages}: results refer to ER-ACE on \tinyimagenet.}
\begin{tabular}{l|cc} 
\toprule
\textbf{Method}                                 & \textbf{FAA}                       & \textbf{FWT}\\
\midrule
\arrayrulecolor{black}
Only Wake               & 4.70  & -0.93\\ 
Wake + REM              & 25.68 & 11.89\\
Wake + NREM             & 27.61 & -0.67\\
Wake + REM + NREM       & 35.68 &  8.60\\
\bottomrule
\end{tabular}
\label{tab:res_ablation}
\end{table}

We further investigate the impact of the size of the dreaming dataset on the results. Figure \ref{fig:dim_dreaming} illustrates how the dreaming stage allows for enhanced performance even when the additional dreaming dataset is reduced by approximately 70\%.

We finally assess the efficiency aspects of WSCL. 
Indeed, the human brain is capable of performing complex tasks with remarkable speed and accuracy, at a relatively low energy cost: cerebral parallel processing architecture, plasticity, and ability to adapt to changing environments are all factors that contribute to its efficiency~\cite{hassabis2017neuroscience,tavanaei2019deep}.
In WSCL, efficiency is encouraged in the wake stage, by letting the model selectively freeze different portions of the network: this is analogous and consistent to cognitive neuroscience evidence that a synchronization of neural activity across different brain regions and changes in the balance between excitation and inhibition enable efficient processing~\cite{tononi2014sleep,marshall2007contribution}.

\begin{figure*}[ht]
  \centering
  \includegraphics[width =0.7\linewidth]{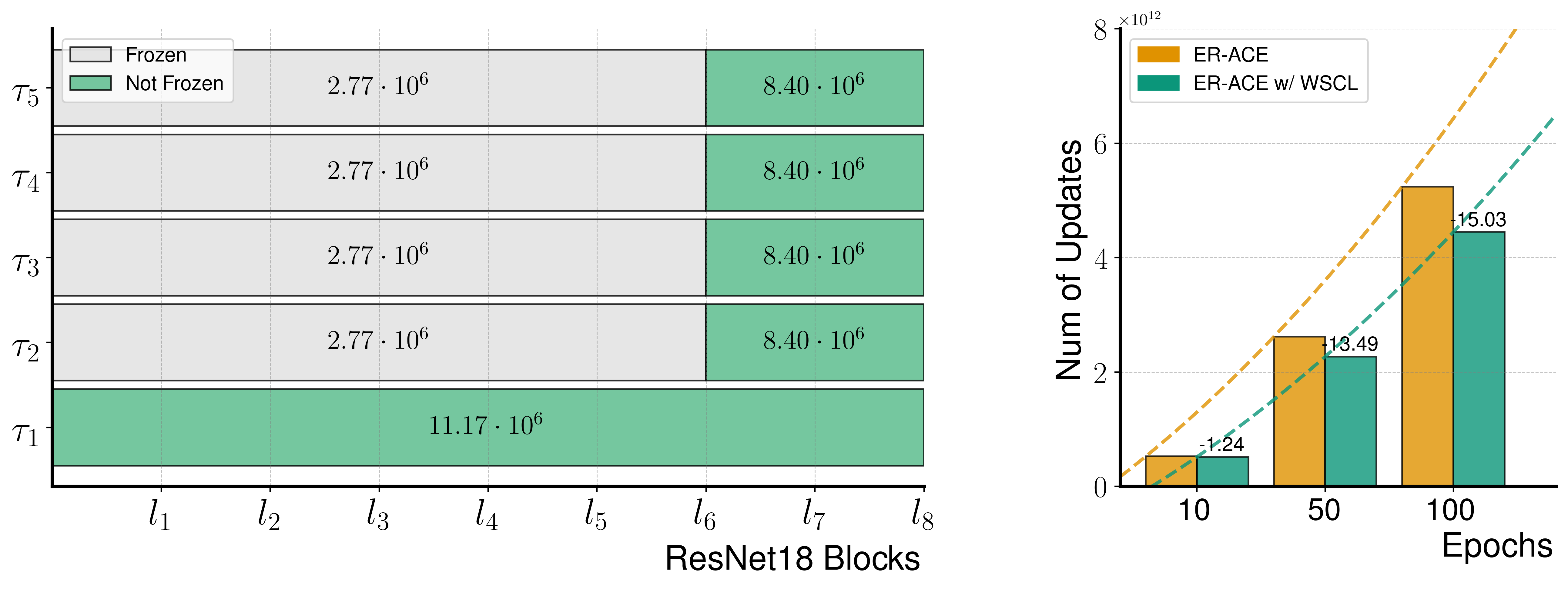}
  \caption{\textbf{WSCL model efficiency}. Left: the most frequent automatically learned freezing scheme (values within bars are number of parameters) during the wake phase for ER-ACE on \tinyimagenet. Right: number of parameter updates for the whole training of ER-ACE with and without WSCL on \tinyimagenet (from 10 epochs to 100 training epochs).}
  \label{fig:efficiency}
\end{figure*}

\begin{figure*}[ht!]
    \centering
    \includegraphics[width =.99\linewidth]{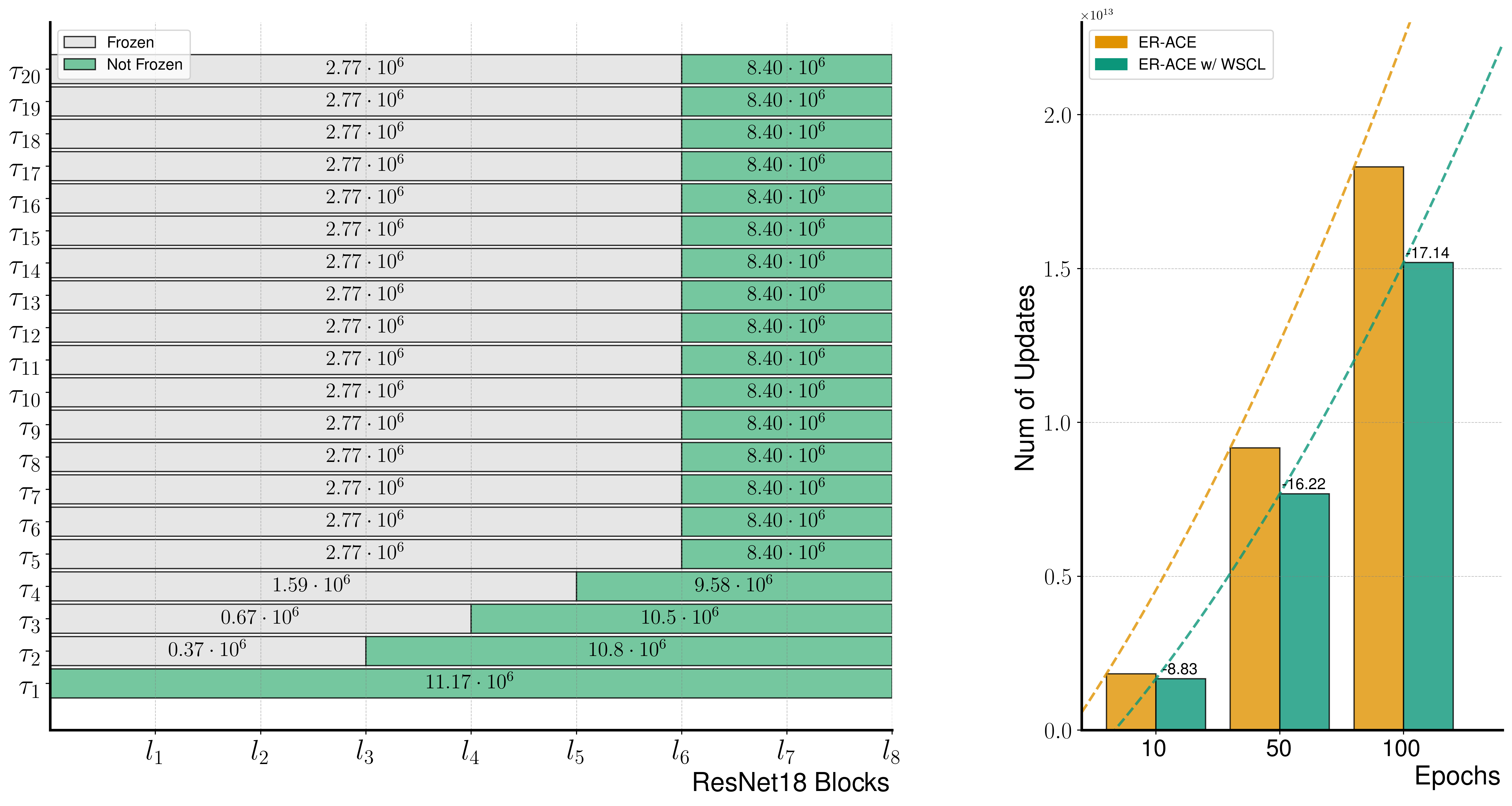}
    \caption{\textbf{WSCL model efficiency}. Left: the most frequent automatically learned freezing scheme (values within bars are number of parameters) during the wake phase for ER-ACE on FG-ImageNet. Right: number of parameter updates for the whole training of ER-ACE with and without WSCL on FG-ImageNet (from 10 epochs to 100 training epochs). The numbers above the green bars represent the improvement in percent points with respect to the baseline alone.}
    \label{fig:frozen_fgimgnet}
\end{figure*}

Fig.~\ref{fig:efficiency} shows the most frequent (over 10 different runs) set of frozen backbone layers at each task, when training ER-ACE with WSCL on \tinyimagenet, as well as the total number of performed parameter updates using the training procedure presented in Sect.~\ref{sec:training}. 
It is important to note that, however, the freezing strategy employed during the wake stage of WSCL depends on the specific continual learning method and the target dataset. Figure \ref{fig:frozen_fgimgnet} illustrates the predominant freezing scheme of ER-ACE when evaluated on the FG-ImageNet dataset, as well as the resulting efficiency. Unlike \tinyimagenet, where ER-ACE typically freezes almost all layers after the completion of the first task, on FG-ImageNet, ER-ACE gradually freezes the layers of its backbone network until Task $\tau_5$. Despite this more gradual freezing strategy, the efficiency gain achieved is approximately 17.14\%, indicating fewer updates compared to the baseline model trained without the wake-sleep strategy in WSCL. 
Thus, WSCL's training procedure reduces the overall number of updates for the entire training of the ResNet-18 model, by a quantity that tends to increase with the number of training epochs (from 2\% to about 17\% less updates), thus confirming the suitability of the wake stage in supporting efficient training.

\section{Conclusion}
The integration of Complementary Learning Systems (CLS) theory and sleep mechanisms in artificial neural networks holds great potential for enhancing continual learning capabilities. Inspired by the interaction between the hippocampus and neocortex in humans, Wake-Sleep Consolidated Learning (WSCL) introduces a sleep phase that mimics off-line brain states during which memory consolidation and synaptic reorganization occur. By leveraging the wake phase for fast adaptation and episodic memory formation, and the sleep phase for memory consolidation and dreaming, WSCL shows superior performance compared to prior work on various benchmarks. Importantly, WSCL achieves positive forward transfer, exhibiting the ability to prepare synapses for future knowledge. These findings highlight the importance of all three stages --- wake, NREM and REM --- in supporting network plasticity and reducing forgetting for improved learning and memory.
Future research will address the advancement of memory and dreaming modeling techniques, which currently rely on conventional rehearsal methods to facilitate memory retention and on the employment of external datasets for generating dream-like experiences. 
With regard to memory modeling, it is essential to delve into more nuanced and dynamic approaches that accurately capture the intricacies of memory formation, storage, and retrieval, by also devising mechanisms to account for memory decay and interference. 
Likewise, for dream modeling, there is an opportunity to push beyond the current reliance on external datasets and explore more sophisticated techniques. This could entail developing generative models capable of simulating dream-like experiences based on the network's existing knowledge and latent representations. By accomplishing this, the model's ability to generate diverse, creative, and contextually relevant dream scenarios can be elevated to a new level of realism.

It is important to acknowledge that, while the pursuit of more realistic memory and dreaming modeling techniques is desirable, their  integration into the WSCL framework is possible thanks to its modular architecture, which provides a solid foundation that can accommodate the inclusion of advanced components dedicated to specific aspects of memory management or sample generation.

\section{Acknowledgements}
This research was supported by the PNRR MUR project PE0000013-FAIR. Matteo Pennisi and Amelia Sorrenti are PhD students enrolled in the National PhD in Artificial Intelligence, XXXVII cycle, course on Health and life sciences, organized by Università Campus Bio-Medico di Roma.


\end{document}